\documentclass[conference]{IEEEtran}
\IEEEoverridecommandlockouts
\usepackage{cite}
\usepackage{amsmath,amssymb,amsfonts}
\usepackage{algorithm}
\usepackage{algorithmic}
\usepackage{graphicx}
\usepackage{textcomp}
\usepackage{xcolor}
\usepackage{booktabs}
\usepackage{threeparttable}
\usepackage[hyphens]{url}
\usepackage{multirow}
\usepackage[hidelinks]{hyperref}
\usepackage{cite}
\def\BibTeX{{\rm B\kern-.05em{\sc i\kern-.025em b}\kern-.08em
    T\kern-.1667em\lower.7ex\hbox{E}\kern-.125emX}}
\begin{document}

\title{Reciprocal Co-Training: Coupling Language Model Encoders and Non-Differentiable Models via Reinforcement Learning}

\author{Yunshuo Tian$^1
$, Akayou Kitessa$^1$, Tanuja Chitnis$^2$, and Yijun Zhao$^1$ \\
  $^1$Department of Computer and Information Science, Fordham University, New York, NY  \\
  $^2${Department of Neurology, Mass General Brigham, Boston, MA
 }  \\
 }


\maketitle
\begin{abstract}
Language models (LMs) and classical machine learning methods offer complementary strengths for predictive modeling, yet their fundamentally different representations and training paradigms hinder effective integration: LMs rely on gradient-based optimization over textual data, whereas models such as Random Forests (RF) employ non-differentiable feature partitioning. This work introduces a reciprocal co-training framework that couples an LM with an RF classifier via reinforcement learning, creating an iterative feedback loop in which each model improves using signals from the other. Tabular data are reformulated into standardized textual representations for the LM, whose embeddings augment the RF feature space, while calibrated RF probability estimates provide feedback signals that guide reinforcement learning updates of the LM. Experiments across three medical datasets, evaluated with both a domain-adapted clinical encoder (ClinicalBERT) and a larger instruction-tuned language model (Qwen2-7B-Instruct), demonstrate consistent performance gains for both model components. Ablation analyses indicate that iterative refinement, hybrid reward design, and dimensionality control jointly contribute to these gains. SHAP analysis further confirms that LM-derived representations are among the most important inputs to the RF predictions. The proposed framework provides a general mechanism that allows incompatible model families to leverage each other’s strengths through bidirectional adaptation.
\end{abstract}

\section{Introduction}

Language models (LMs) have emerged as powerful pretrained models capable of learning rich representations from large-scale corpora \cite{vaswani2017attention,brown2020language,touvron2023llama}. While such pretrained models demonstrate strong generalization across many tasks, effective performance on specialized applications often requires adaptation or additional pretraining with domain-specific data \cite{gururangan2020don}. Recent work explores applying LMs to feature-based prediction problems by reformulating variables into textual descriptions that allow models to leverage pretrained knowledge \cite{dinh2022tabLM}. Nevertheless, empirical studies have shown that classical machine learning methods often perform as well as or better than neural approaches on many datasets, particularly when training samples are limited \cite{grinsztajn2022why}.

Among classical approaches, tree-based ensembles such as Random Forests (RF) often provide strong baselines due to their robustness and reliable performance across diverse datasets \cite{breiman2001random,grinsztajn2022why}. These models partition feature space through ensembles of decision trees and often perform well when data are heterogeneous or sample sizes are moderate. In contrast, LMs learn contextual representations that capture complex relationships once inputs appear in natural language form. Despite these complementary strengths, existing approaches typically apply LMs and classical models independently. As a result, neither model benefits from the strengths of the other.

Existing hybrid approaches typically transfer information in one direction, either by using LM embeddings as inputs to a downstream classifier \cite{dinh2022tabLM,ijcai2025p687} or by combining independently trained models through ensembling \cite{wolpert1992stacked,kablan2023evaluation,alzubaidi2023stacking}. Such approaches do not allow the underlying representations and decision boundaries to adapt to each other. This limitation motivates a framework in which both models adapt through iterative feedback, allowing representations and decision boundaries to be refined jointly during training.

This study proposes a reciprocal co-training framework (RCT) that links a language model encoder with a Random Forest classifier through alternating optimization. Because tree ensembles are inherently non-differentiable while language model encoders rely on gradient-based learning, direct end-to-end training is not feasible. Reinforcement learning serves as a bridge between the two models: RF predictions are converted into reward signals that guide updates to the language model representation, while the resulting embeddings augment the RF feature space. In contrast to conventional pipelines where one component remains fixed, the proposed framework allows both models to adapt to each other during training and iteratively refine representations and predictions.

The framework is evaluated on predicting three-year relapse at the first clinical visit in multiple sclerosis (MS), a chronic neurological disease characterized by heterogeneous progression \cite{madill2024prediction}. Early relapse prediction has significant clinical value through support for timely intervention and treatment planning. Two additional public medical datasets assess generalization beyond the MS domain. Across all datasets, RCT consistently improves predictive performance for both the language model and RF components. Feature-importance analysis further confirms that LM-derived representations play a substantial role in the final RF model. Experiments with both ClinicalBERT (CBERT)\cite{alsentzer2019clinicalbert} and Qwen2-7B-Instruct (Qwen2) \cite{yang2024qwen2} demonstrate that the framework generalizes across domain-specific and instruction-tuned language model architectures. These results provide evidence that reinforcement learning can bridge gradient-based language models with a non-differentiable Random Forest model and allow heterogeneous learners to benefit from bidirectional feedback. Implementation of the framework is provided under \cite{github} to facilitate reproducibility.

\section{Related Work}
Prior work relevant to this study can be largely organized into methodological advances in LM-based hybrid learning and clinical applications of language models for prediction from electronic health record (EHR) data.

\subsection{Language Models and Hybrid Learning}

Recent research has explored adapting LMs to feature-based prediction tasks. Beyond text generation, several approaches attach task-specific classification heads to transformer embeddings and employ parameter-efficient fine-tuning methods such as LoRA \cite{hu2021lora}. Other work reformulates feature vectors as prompts to leverage pretrained language model priors \cite{dinh2022tabLM}. More specialized frameworks align latent transformer representations with downstream encoders rather than relying on textual outputs. For example, Latte \cite{ijcai2025p687} extracts hidden-state representations and transfers latent knowledge for few-shot learning on structured datasets, mitigating hallucination and latency associated with text-based pipelines. Related work has also explored transformer architectures designed specifically for this scenario, including TabPFN \cite{hollmann2023tabpfn} and other deep models evaluated in recent benchmarks \cite{gorishniy2021revisiting}. These approaches span a spectrum from domain-adapted encoders pretrained on specialized corpora to general-purpose instruction-tuned models, but they primarily rely on direct supervised adaptation of the language model.

Efforts to combine foundation models with classical predictors remain comparatively limited. Hybrid modeling strategies such as stacked generalization \cite{wolpert1992stacked} combine heterogeneous learners but typically train components independently without iterative mutual adaptation.

Reinforcement learning has played a central role in facilitating the incorporation of external feedback signals into large language models \cite{ouyang2022training} and in stabilizing policy updates through methods such as Proximal Policy Optimization (PPO) \cite{schulman2017ppo}. However, prior work generally optimizes a single differentiable model. The use of reinforcement learning to couple a non-differentiable Random Forest with a transformer through alternating updates has received limited attention.

\subsection{Language Models and Classical Approaches in Clinical Prediction}

In the clinical domain, large language models have increasingly been applied to electronic health record (EHR) data for risk prediction and disease monitoring. Domain-adapted transformer models such as ClinicalBERT, trained on clinical notes, provide contextual representations that improve performance on downstream clinical prediction tasks \cite{alsentzer2019clinicalbert}. Fine-tuning ClinicalBERT on EHR data has demonstrated improvements over traditional baselines for tasks such as hospital readmission prediction and other clinical outcome modeling \cite{huang2019clinicalbert}. In multiple sclerosis (MS), transformer-based approaches have been used to predict Expanded Disability Status Scale outcomes and relapse risk \cite{zhan2023precision}. Other studies integrate natural language processing with classical models to extract disease activity signals from unstructured clinical documentation \cite{chang2022detecting}.

Classical ensemble models such as Random Forests \cite{breiman2001random} and XGBoost \cite{chen2016xgboost} are frequently used as reference baselines in clinical modeling studies. Prior work that combines neural networks with ensemble models typically trains the components independently through stacking or hybrid ensembles \cite{kablan2023evaluation,alzubaidi2023stacking}, or transfers learned representations from neural networks to classical classifiers \cite{akbar2022covid}. These approaches rely on either independent training or one-directional representation transfer, whereas the framework considered in this study allows the two model families to adapt to each other during training through iterative feedback.

\section{Datasets and Preprocessing}
This section describes the datasets employed in this study and model-specific preprocessing steps.

\begin{figure}[!t]
\centering
\small
\fbox{%
\begin{minipage}{0.9\linewidth}
\noindent\textbf{Input} \\[0.5em]
Age: 34.2 years \\
Sex: Male \\
Race: Black or African American \\
Disease Duration at First Visit: 1.3 years \\
Disease Category: Relapsing-Remitting MS \\
EDSS Score: 1.5 \\
Total Relapses Before FV: 2 \\
\hspace*{2cm} $\vdots$ \\
New T2 Lesion in Past Year: No \\
New Gad Lesion in Past Year: Yes \\

\noindent\textbf{Future Relapse Label:} \textbf{No}
\end{minipage}
}
\caption{Sample EHR-to-Text Conversion}
 \label{fig:ehr}
 \end{figure}

\begin{figure*}
    \centering
\includegraphics[ trim = 100 120 100 0, clip, scale=0.4]
{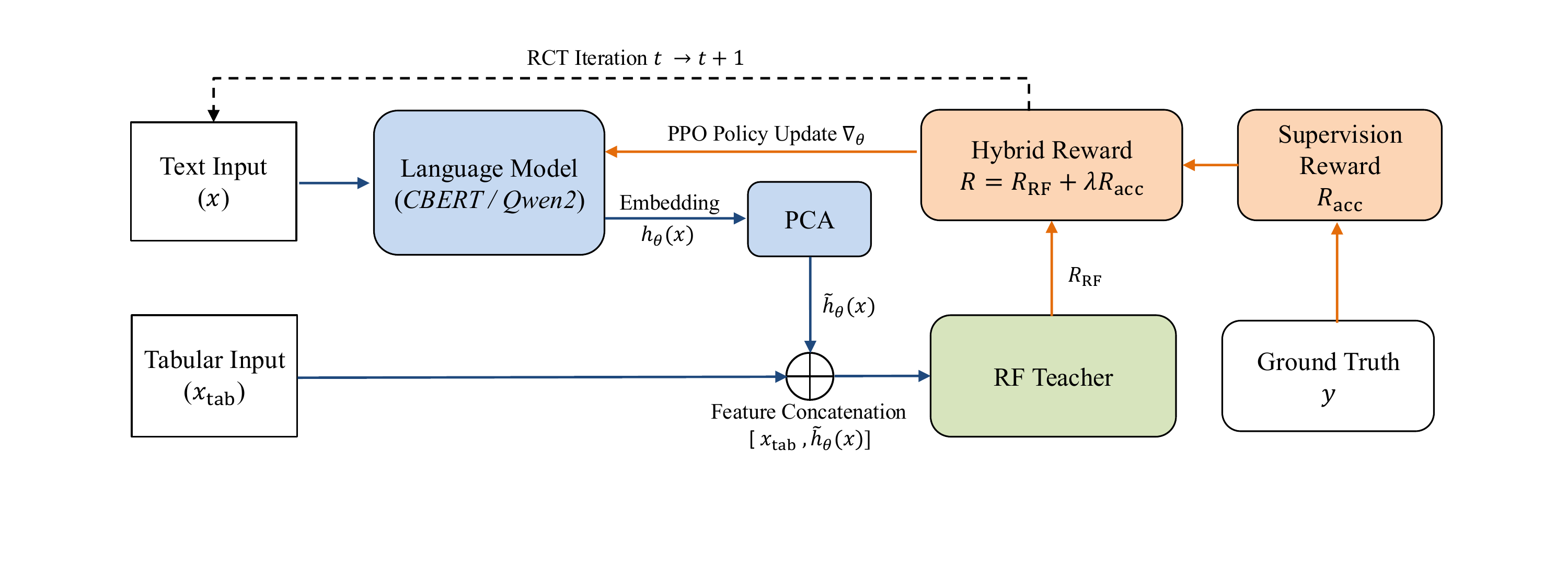}
    \caption{Overview of the Reciprocal Co-Training (RCT) framework. Blue arrows indicate forward data flow, while orange arrows indicate reward feedback for PPO updates. The dashed loop denotes one RCT iteration.}
    \label{fig:framework}
\end{figure*}

\subsection{Datasets}

The primary dataset consists of 2,192 patients with clinically isolated syndrome (CIS) or relapsing-onset MS enrolled in the Comprehensive Longitudinal Investigation of Multiple Sclerosis at Brigham and Women’s Hospital (CLIMB) study \cite{gauthier2006model}. This study was approved by the Mass General Brigham Institutional Review Board. The need for individual patient consent was waived as part of a secondary use protocol due to the low-risk nature and prior consent for the biorepository. 

The task is to determine whether a patient is likely to experience a clinical relapse within three years following the first clinical visit (FV). 
Eligibility criteria required patients to be $\geq$18 years old at FV with a diagnosis of CIS or relapsing-onset MS. A minimum of three years of follow-up was required to ensure adequate outcome observation. Patients with progressive MS at baseline or prior exposure to high-efficacy or long-acting therapies before FV were excluded to avoid confounding early disease activity assessment. Predictor variables recorded at FV include demographic factors (age, sex, race), disease duration, Expanded Disability Status Scale (EDSS), MS subtype, relapse history, MRI activity indicators, smoking history, family history, and treatment-related variables. The outcome is defined as the occurrence of a clinician-adjudicated relapse between 30 days and three years after FV.

Two widely used benchmarks from the UCI ML Repository \cite{asuncion2007uci} are used to evaluate generalization beyond the MS prediction task. The Wisconsin Diagnostic Breast Cancer dataset contains 569 samples with 30 continuous features for malignant versus benign tumor classification \cite{street1993nuclear}. The BRFSS Diabetes dataset includes 70{,}692 samples with 8 clinical variables for diabetes versus non-diabetes prediction \cite{cdc2015brfss}. They provide moderate-sample datasets distinct from the MS cohort while preserving comparable tabular input characteristics.

For each dataset, samples are randomly partitioned into training and test sets using an 80/20 split. Model training and hyperparameter selection are performed on the training portion, while final performance is evaluated on the held-out test set.

\subsection{EHR-to-Text Reformulation for LM}
\label{sec:reformulation}
Each EHR record is reformulated into a standardized patient-card representation as a sequence of attribute-value pairs. The resulting representation is compatible with transformer-based language models and preserves the full information content of the original feature matrix. An illustrative MS patient card is presented in Figure \ref{fig:ehr}. The transformation is deterministic and introduces no additional synthetic narrative content. The same encoding template is applied consistently across all records of the proprietary and public datasets.

\subsection{Preprocessing for Random Forest}

For the RF approach, categorical variables are one-hot encoded, and continuous variables are retained in their original scale. Features with more than 50\% missing values are removed during preprocessing. Class imbalance in the MS dataset (positive rate $\approx$ 36\%) is addressed through cost-sensitive learning using \texttt{class\_weight=``balanced\_subsample"}, which reweights classes inversely proportional to their frequency. For the standalone RF baseline, hyperparameters were selected using
grid search with 10-fold cross-validation on the training data.
The resulting configuration was then fixed for all subsequent
experiments and iterations of the RCT framework. During PPO training of the language model, positive samples are oversampled with a weight of 1.5 to increase their sampling probability. An asymmetric reward penalizes false negatives more strongly than false positives (FN=$-1.5$ vs.\ FP=$-0.2$), encouraging sensitivity to the minority class. 

Within the iterative co-training process, the RF model is trained on an augmented feature space incorporating LM-derived embeddings.

\section{Method}
Figure \ref{fig:framework} illustrates the RCT framework. Structured inputs are reformulated into text (Section \ref{sec:reformulation}) and processed by a language model to produce contextual embeddings. These embeddings are reduced in dimension and concatenated with the original features to train an RF classifier. The RF generates calibrated probability estimates that, together with ground-truth supervision, define a hybrid reward used to update the language model via reinforcement learning. The process then repeats until convergence. Conceptually, the RF provides feedback on the LM’s predictions, while the LM generates contextual embeddings that augment the feature space used by the RF. The two models exchange information through alternating optimization. We next formalize this framework by defining the problem setup, followed by the LM update, the RF update, and the overall training procedure.

\subsection{Problem Formulation}

Consider a binary prediction problem with tabular input 
$x_{\text{tab}} \in \mathbb{R}^d$ and label $y \in \{0,1\}$. 
This representation is reformulated into a textual prompt $x$ using the patient-card representation described in Section \ref{sec:reformulation}.

Let $f_\theta$ denote a language model parameterized by $\theta$. A binary classification head is attached to the final hidden representation. 
In this work, we evaluate two backbone architectures: CBERT \cite{alsentzer2019clinicalbert}, a BERT encoder pretrained on clinical notes from MIMIC-III \cite{johnson2016mimiciii}, and Qwen2 \cite{yang2024qwen2}, a decoder-only instruction-tuned large language model. Both models are fine-tuned using LoRA \cite{hu2021lora}, while the pretrained model parameters remain frozen. Only the LoRA adapters, classification head, and value head are updated during PPO training. The model defines a stochastic policy
\begin{equation}
\pi_\theta(a \mid x), \quad a \in \{0,1\},
\end{equation}
where $a$ is sampled from the softmax distribution over logits produced by the classification head during reinforcement learning, while inference uses the predicted class probability.

Let $g_\phi$ denote a RF classifier parameterized by $\phi$, producing probability estimates
\begin{equation}
p_\phi(y = 1 \mid x_{\text{RF}}).
\end{equation}
The two model families capture different aspects of the data: RFs operate directly on structured features and provide stable probabilistic decision boundaries, whereas LMs capture complex feature interactions through contextual modeling. 
The goal is to optimize $\theta$ and $\phi$ through an alternating procedure in which the two models iteratively refine each other.

\subsection{Alternating Co-Training Framework}

Let $(\theta^{(t)}, \phi^{(t)})$ denote model parameters at outer iteration $t$. 
Training alternates between updating the LM while holding the RF fixed, and updating the RF while holding the LM fixed:

\begin{align}
\theta^{(t+1)} &\approx \arg\max_\theta \, J(\theta; \phi^{(t)}), \\
\phi^{(t+1)}   &= \arg\min_\phi \, L_{\text{RF}}(\phi; \theta^{(t+1)}).
\end{align}
where $J(\theta; \phi^{(t)})$ denotes the PPO objective defined under fixed RF parameters, and 
$L_{\text{RF}}$ denotes supervised empirical risk on the augmented feature representation produced by the updated LM.

\subsection{LM Update via Hybrid Reward}

Because the RF is non-differentiable, its outputs cannot be
backpropagated through the language model. During the LM
update phase, the RF parameters $\phi^{(t)}$ are held fixed, and
RF outputs are treated as scalar reward signals without gradient
flow. For input $x$, the LM samples an action
$a \sim \pi_\theta(\cdot \mid x)$, where $a \in \{0,1\}$.

Let $p_\phi(y=1 \mid x_{\text{RF}})$ denote the RF-estimated
probability for the positive class. For a sampled action $a$, we
define the RF action value as
\begin{equation}
Q_\phi(x,a)=
\begin{cases}
p_\phi(y=1 \mid x_{\text{RF}}), & a=1,\\
1-p_\phi(y=1 \mid x_{\text{RF}}), & a=0.
\end{cases}
\end{equation}
The RF reward is based on a centered action advantage. Specifically,
\begin{equation}
V_\phi(x)
=
\frac{1}{2}
\left[
p_\phi(y=1 \mid x_{\text{RF}})
+
1-p_\phi(y=1 \mid x_{\text{RF}})
\right],
\end{equation}
and
\begin{equation}
A_{\text{RF}}(x,a)=Q_\phi(x,a)-V_\phi(x).
\end{equation}
For binary classification, $V_\phi(x)=0.5$. We retain the explicit form to emphasize that the RF reward measures the sampled action relative to the average value of the two possible actions. The RF component of the reward is then
clipped and scaled:
\begin{equation}
R_{\text{RF}}(x,a)
=
s \cdot
\operatorname{clip}
\left(A_{\text{RF}}(x,a), -c, c\right).
\end{equation}
In all experiments, $s=2.0$ and $c=0.45$.

The RF component depends only on the sampled action and the
RF probability estimate; label information enters through the
supervised reward defined below.

Ground-truth supervision is incorporated through an accuracy
reward,
\begin{equation}
R_{\text{acc}}(a,y)=
\begin{cases}
1.0, & a=1,\; y=1,\\
0.2, & a=0,\; y=0,\\
-1.5, & a=0,\; y=1,\\
-0.2, & a=1,\; y=0.
\end{cases}
\end{equation}
The supervised component assigns the strongest penalty to false negatives and a smaller penalty to false positives, reflecting the
clinical emphasis on sensitivity in risk prediction tasks. The final hybrid reward is additive after RF centering, clipping, and
scaling:
\begin{equation}
R(x,a,y)
=
R_{\text{RF}}(x,a)
+
\lambda R_{\text{acc}}(a,y),
\end{equation}
with $\lambda=0.5$ in all experiments.

The LM policy is updated by minimizing the PPO loss,
\begin{equation}
\begin{aligned}
L_{\text{PPO}}(\theta)
=&-\mathbb{E}\Big[
\min \big(
r_t(\theta) A_t, \\
&\quad
\operatorname{clip}
\big(r_t(\theta), 1-\epsilon, 1+\epsilon\big) A_t
\big)
\Big] \\
&+ c_1 L_{\text{value}} - c_2 H(\pi_\theta).
\end{aligned}
\end{equation}
where
\begin{equation}
r_t(\theta)=
\frac{\pi_\theta(a \mid x)}
{\pi_{\theta_{\text{old}}}(a \mid x)}.
\end{equation}

\noindent Here, $A_t$ is computed from the hybrid reward and normalized
within mini-batches before PPO updates. We use $\epsilon=0.2$,
$c_1=0.5$, and $c_2=0.05$. During this phase, only the LoRA
adapters, classification head, and value head are updated; the RF
classifier remains fixed until the subsequent RF refresh step.

\subsection{RF Update with LM Embeddings}

After the LM update, its parameters $\theta^{(t+1)}$ are held fixed. For each input $x$, an embedding $h_\theta(x)$ is extracted based on the underlying architecture. For CBERT, $h_\theta(x)$ corresponds to the final hidden state of the [CLS] token, whereas for Qwen2 it corresponds to the final hidden representation of the last token. To control dimensionality, reduce computational cost, and mitigate overfitting, PCA is applied to the LM embeddings before RF training. CBERT embeddings (768 dimensions) are reduced to five principal components, whereas Qwen2 embeddings (4096 dimensions) are reduced to 50 principal components to accommodate the larger embedding space.

\begin{equation}
\tilde{h}_\theta(x) = \text{PCA}(h_\theta(x)).
\end{equation}

The RF feature vector is defined as 
\begin{equation}
x_{\text{RF}} = [x_{\text{tab}}, \tilde{h}_\theta(x)].
\end{equation}
and the RF parameters are then updated via supervised empirical risk minimization:
\begin{equation}
\phi^{(t+1)} = \arg\min_\phi L_{\text{RF}}(\phi; x_{\text{RF}}, y).
\end{equation}

\begin{algorithm}[!t]
\caption{Iterative LM--RF Co-Training}
\label{alg:cotrain}
 \small
\begin{algorithmic}[1]
\STATE Initialize LM parameters $\theta^{(0)}$
\STATE Train initial RF $\phi^{(0)}$ on $x_{\text{tab}}$
\FOR{$t = 0,1,\dots,T-1$}
    \STATE \textbf{LM update:} Fix RF parameters $\phi^{(t)}$
    \FOR{$k = 1,\dots,K$}
        \STATE Sample $a \sim \pi_{\theta}(a \mid x)$
        \STATE Compute RF action value $Q_{\phi^{(t)}}(x,a)$
        \STATE Compute centered RF advantage $A_{\text{RF}}(x,a)$
        \STATE Compute clipped and scaled reward $R_{\text{RF}}(x,a)$
        \STATE Compute hybrid reward $R(x,a,y)$
        \STATE Update $\theta$ using PPO
    \ENDFOR
    \STATE Obtain updated LM parameters $\theta^{(t+1)}$
    \STATE \textbf{RF update:} Fix $\theta^{(t+1)}$
    \STATE Extract embeddings $h_{\theta^{(t+1)}}(x)$
    \STATE Reduce embeddings with PCA
    \STATE Form augmented features $[x_{\text{tab}}, \tilde{h}_{\theta^{(t+1)}}(x)]$
    \STATE Retrain RF to obtain $\phi^{(t+1)}$
    \STATE Evaluate validation ROC-AUC
    \IF{no improvement for $P$ consecutive iterations}
        \STATE \textbf{break}
    \ENDIF
\ENDFOR
\STATE \textbf{return} $(\theta^*,\phi^*)$
\end{algorithmic}
\end{algorithm}

\subsection{Training Procedure}

The RCT procedure is summarized in Algorithm \ref{alg:cotrain}. 
Each outer iteration consists of an LM update phase using PPO under fixed RF parameters, followed by an RF retraining phase using the updated LM embeddings. 
Early stopping (patience $=5$) is applied on the validation ROC-AUC scores. All reward signals are computed on the training data, while evaluation uses the test data. Code and experimental scripts are provided under \cite{github}.

\setlength{\tabcolsep}{8pt}
\renewcommand{\arraystretch}{1.15}
\begin{table*}[!t]
\caption{Predictive Performance of the RCT Framework Compared with Standalone Baselines on Three Datasets}
\label{tab:main}
\centering
\small
\begin{threeparttable}
\begin{tabular}{llllccccc}
\hline
Category & Model & ROC-AUC & PR-AUC & F1 & Accuracy & Precision & Specificity & Recall \\
\hline

\multicolumn{9}{c}{MS Dataset ($n=2192$)} \\ \cline{2-9}

Baselines & DT    & 0.635 & 0.443 & 0.558 & 0.529 & 0.419 & 0.360 & 0.837 \\
          & RF    & 0.686 & 0.505 & 0.576 & 0.578 & 0.449 & 0.453 & 0.806 \\
          & XGB   & 0.685 & 0.504 & 0.581 & 0.585 & 0.455 & 0.464 & 0.806 \\
          & CBERT & 0.672 & 0.525 & 0.595 & 0.589 & 0.472 & 0.461 & 0.805 \\
          & Qwen2 & 0.660 & 0.536 & 0.563 & 0.532 & 0.433 & 0.370 & 0.805 \\ \cline{2-9}

RCT (CBERT) & RF
& \textbf{0.721} (+0.035)
& \textbf{0.607} (+0.102)
& 0.599 & 0.597 & 0.477 & 0.473 & 0.805 \\
& CBERT
& \textbf{0.700} (+0.028)
& \textbf{0.539} (+0.014)
& 0.600 & 0.599 & 0.478 & 0.476 & 0.805 \\

RCT (Qwen2) & RF
& \textbf{0.720} (+0.034)
& \textbf{0.582} (+0.077)
& 0.614 & 0.622 & 0.496 & 0.513 & 0.805 \\
& Qwen2
& \textbf{0.705} (+0.045)
& \textbf{0.573} (+0.037)
& 0.587 & 0.576 & 0.462 & 0.440 & 0.805 \\
\hline

\multicolumn{9}{c}{Breast Cancer Dataset ($n=569$)} \\ \cline{2-9}

Baselines & DT    & 0.951 & 0.952 & 0.906 & 0.888 & 0.957 & 0.933 & 0.861 \\
          & RF    & 0.987 & 0.989 & 0.888 & 0.871 & 0.987 & 0.981 & 0.807 \\
          & XGB   & 0.990 & 0.993 & 0.888 & 0.872 & 0.990 & 0.986 & 0.806 \\
          & CBERT & 0.736 & 0.646 & 0.602 & 0.605 & 0.479 & 0.486 & 0.810 \\
          & Qwen2 & 0.970 & 0.963 & 0.891 & 0.927 & 0.990 & 0.995 & 0.810 \\ \cline{2-9}

RCT (CBERT) & RF
& \textbf{0.999} (+0.012)
& \textbf{0.998} (+0.009)
& 0.988 & 0.991 & 1.000 & 1.000 & 0.976 \\
& CBERT
& \textbf{0.999} (+0.263)
& \textbf{0.999} (+0.353)
& 0.988 & 0.991 & 1.000 & 1.000 & 0.976 \\

RCT (Qwen2) & RF
& \textbf{0.998} (+0.011)
& \textbf{0.997} (+0.008)
& 0.895 & 0.930 & 1.000 & 1.000 & 0.810 \\
& Qwen2
& \textbf{0.985} (+0.015)
& \textbf{0.978} (+0.015)
& 0.895 & 0.930 & 1.000 & 1.000 & 0.810 \\
\hline

\multicolumn{9}{c}{Diabetes Dataset ($n=70,692$)} \\ \cline{2-9}

Baselines & DT    & 0.798 & 0.403 & 0.431 & 0.663 & 0.294 & 0.636 & 0.808 \\
          & RF    & 0.816 & 0.437 & 0.457 & 0.701 & 0.320 & 0.682 & 0.800 \\
          & XGB   & 0.823 & 0.459 & 0.465 & 0.710 & 0.328 & 0.693 & 0.800 \\
          & CBERT & 0.816 & 0.790 & 0.755 & 0.741 & 0.715 & 0.681 & 0.800 \\
          & Qwen2 & 0.811 & 0.766 & 0.755 & 0.741 & 0.715 & 0.682 & 0.800 \\ \cline{2-9}

RCT (CBERT) & RF
& \textbf{0.829} (+0.013)
& \textbf{0.804} (+0.367)
& 0.761 & 0.749 & 0.725 & 0.697 & 0.801 \\
& CBERT
& \textbf{0.826} (+0.010)
& \textbf{0.797} (+0.007)
& 0.760 & 0.747 & 0.724 & 0.694 & 0.801 \\

RCT (Qwen2) & RF
& \textbf{0.823} (+0.007)
& \textbf{0.798} (+0.361)
& 0.758 & 0.744 & 0.719 & 0.688 & 0.800 \\
& Qwen2
& \textbf{0.813} (+0.002)
& \textbf{0.773} (+0.007)
& 0.754 & 0.739 & 0.713 & 0.679 & 0.800 \\
\hline
\end{tabular}

\begin{tablenotes}[flushleft]
\item Parenthetical values denote absolute changes relative to the corresponding standalone model: RF rows are compared with the standalone RF, and CBERT/Qwen2 rows are compared with the corresponding standalone LM. Boldface highlights improvement. All metrics are reported at matched sensitivity $\approx 0.80$ (if possible).\\
\end{tablenotes}
\end{threeparttable}
\end{table*}

\section{Results}

This section evaluates the proposed framework on three datasets of varying size and difficulty. 
We first compare model 
performance against independently trained baselines, then analyze the training dynamics of the iterative procedure, and finally assess the contribution of individual design components through controlled ablation experiments.

To support fair comparison across models, classification thresholds are calibrated post-hoc to achieve comparable recall at approximately $0.80$ for each model. Under this setting, differences in specificity and precision primarily reflect variation in false-positive control while maintaining similar relapse detection rates.

\begin{figure*}[!t]
\vspace{2mm}
    \centering

\includegraphics[ trim = 120 550 300 132, clip, scale=0.45]
{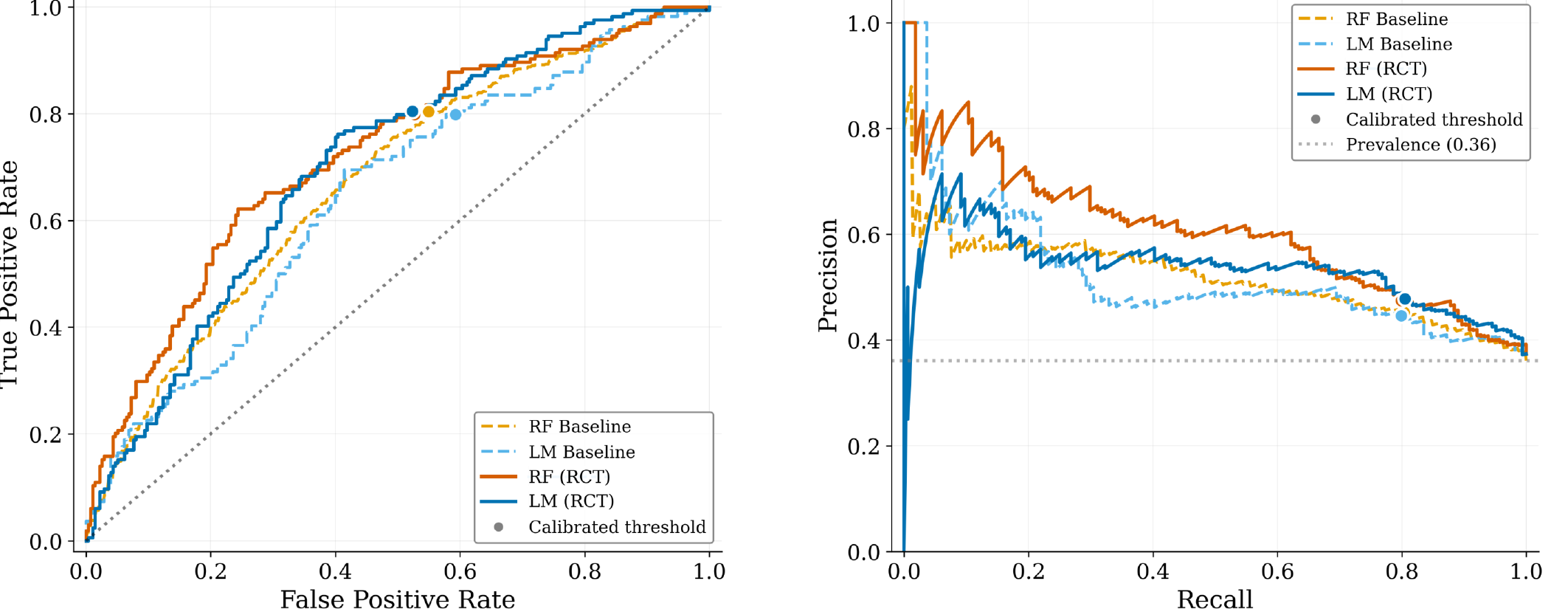}

\caption{\small ROC (left) and precision–recall (right) curves for relapse prediction on the MS dataset using CBERT. The proposed RCT model consistently dominates the baselines across most thresholds.}
\label{fig:curves} 
\end{figure*}

\begin{figure*}[!t]
  \centering
\includegraphics[ trim = 150 515 360 280, clip, scale=0.54]
{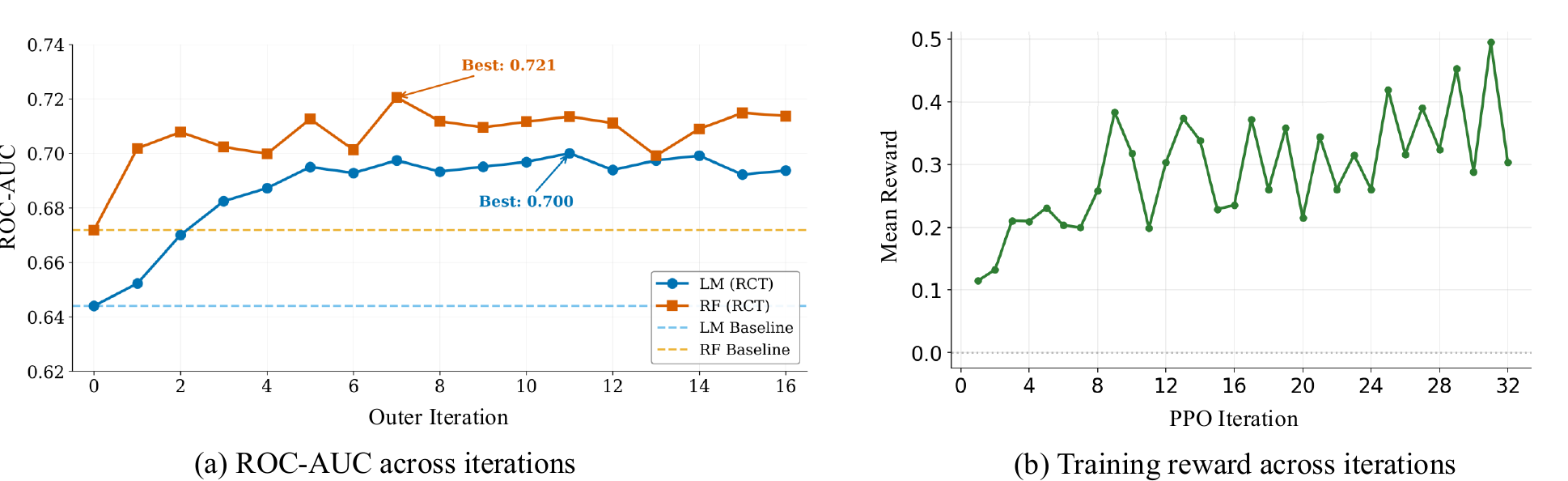}
\caption{\small Training dynamics of the RCT process on the MS-Relapse dataset using CBERT. 
(a) ROC-AUC of the RF and LM across outer iterations of alternating optimization. 
(b) Mean PPO reward during LM updates.}
\label{fig:dynamics} 
\end{figure*}

\subsection{Predictive Performance}

Table \ref{tab:main} reports predictive performance on the three datasets. Results are reported at a matched sensitivity of approximately 0.80 to facilitate comparison across methods. We evaluate both components of the reciprocal framework: the language model after reciprocal co-training and the RF retrained with LM-derived embeddings. Standalone baseline results were averaged over three independent runs. ROC-AUC standard deviations were small across all datasets, and the relative ranking of the baseline methods remained unchanged across runs. RCT results are reported from a single run due to computational constraints.

On the MS-Relapse dataset, reciprocal co-training improves both the language model and RF performance. For CBERT, RF ROC-AUC increases from 0.686 to 0.721 (+0.035), while the language model increases from 0.672 to 0.700 (+0.028). For Qwen2, RF ROC-AUC increases from 0.686 to 0.720 (+0.034), while the language model increases from 0.660 to 0.705 (+0.045). PR-AUC follows the same pattern. The CBERT-based RF increases from 0.505 to 0.607 (+0.102), and the Qwen2-based RF increases from 0.505 to 0.582 (+0.077). Precision and specificity also increase with minimal impact on recall, and these gains are consistent across both encoder types and across both model components. Therefore, language-model representations likely provide complementary information for RF prediction. 

Figure~\ref{fig:curves} presents the ROC and precision–recall curves for CBERT on the MS-Relapse dataset. The RCT models dominate their standalone counterparts across a wide range of classification thresholds. This global improvement supports that the gains are not limited to a single operating point. Qwen2 exhibits similar behavior, with ROC-AUC and PR-AUC values comparable to those of CBERT (Table \ref{tab:main}).

For the Breast Cancer dataset, both CBERT and Qwen2 achieve near-ceiling performance after reciprocal co-training. WDBC serves as a controlled benchmark with high class separability. The weak CBERT baseline is expected because CBERT was pretrained on clinical narrative text, whereas WDBC consists of numerical cell-nucleus measurements that, when serialized, do not resemble clinical language. Qwen2, pretrained on more extensive and diverse text corpora, adapts more readily to serialized tabular descriptions. Consequently, Qwen2 starts from a strong standalone baseline (ROC-AUC 0.970) and reaches 0.985 (+0.015), while the associated RF increases from 0.987 to 0.998 (+0.011). CBERT starts from a lower baseline (ROC-AUC 0.736), which leaves more room for improvement. Reciprocal co-training raises CBERT performance to 0.999 (+0.263), while the associated RF reaches 0.999 (+0.012). PR-AUC exhibits the same pattern. The larger CBERT gains likely reflect its weaker standalone performance on this dataset rather than an advantage of the reciprocal framework for one LM over the other.

The Diabetes dataset presents a more challenging and highly imbalanced prediction task. Improvements in ROC-AUC are smaller than those observed on the MS dataset, likely because both model families already achieve strong standalone performance. Although ROC-AUC improvements are modest, substantially larger gains are observed in PR-AUC, which indicates improved ranking of minority-class instances under class imbalance. For CBERT, RF ROC-AUC increases from 0.816 to 0.829 (+0.013), while the language model increases from 0.816 to 0.826 (+0.010). For Qwen2, RF ROC-AUC increases from 0.816 to 0.823 (+0.007), while the language model increases from 0.811 to 0.813 (+0.002). The largest effect appears in PR-AUC. The RF baseline achieves a PR-AUC of 0.437, whereas the CBERT-augmented RF reaches 0.804 and the Qwen2-augmented RF reaches 0.798. The PR-AUC improvement suggests that LM-derived representations improve the model's ability to rank minority-class instances, particularly in the precision-recall regime where RF baselines are weaker. 

Across the three benchmarks, both language models benefit from RCT. On the Breast Cancer dataset, the weaker CBERT baseline yields larger performance gains after training, whereas Qwen2 starts from a stronger baseline and therefore exhibits smaller improvements. The RF component also outperforms the standalone RF, with the largest gains observed on the MS-Relapse and Diabetes datasets.

\begin{table}[t]
\centering
\small
\begin{threeparttable}
\caption{Convergence Summary Across Datasets and Language Models}
\label{tab:convergence}

\begin{tabular}{llcccc}
\toprule
\textbf{Dataset} & \textbf{Model}
& \textbf{LM}
& \textbf{Iter}
& \textbf{RF}
& \textbf{Iter} \\
\midrule

\multirow{2}{*}{MS}
& CBERT
& 0.700 & 11
& 0.721 & 7 \\
& Qwen2
& 0.705 & 12
& 0.720 & 8 \\
\midrule

\multirow{2}{*}{WDBC}
& CBERT
& 0.999 & 18
& 0.999 & 18 \\
& Qwen2
& 0.985 & 1
& 0.998 & 1 \\
\midrule

\multirow{2}{*}{BRFSS}
& CBERT
& 0.826 & 5
& 0.829 & 5 \\
& Qwen2
& 0.813 & 6
& 0.823 & 1 \\
\bottomrule
\end{tabular}

\begin{tablenotes}[flushleft]
\footnotesize
\item LM and RF denote the best test ROC-AUC achieved by the language model and Random Forest components, respectively. Iter specifies the outer iteration at which the best value was first attained. Iteration 0 corresponds to the baseline model prior to reciprocal co-training. MS = MS-Relapse, WDBC = Breast Cancer Wisconsin Diagnostic, and BRFSS = Diabetes BRFSS.
\end{tablenotes}
\end{threeparttable}
\end{table}

\subsection{Training Dynamics Analysis}

Figure \ref{fig:dynamics} and Table \ref{tab:convergence} summarize the training dynamics of the RCT process. Across all datasets and models, both the LM and RF components improve rapidly during the early iterations before gradually stabilizing. The performance trajectories remain stable throughout training and exhibit no large performance oscillations, which provides supporting evidence that the reciprocal exchange of information between the two model families produces a reliable optimization process.

On the MS-Relapse dataset, performance gains are distributed across multiple reciprocal updates. Figure~\ref{fig:dynamics}(a) illustrates the ROC-AUC trajectory for CBERT. Both the language model and RF surpass their standalone baselines within the early iterations, consistent with the benefit of reciprocal feedback during training. The RF reaches its best ROC-AUC of 0.721 at iteration 7, while the language model reaches its best ROC-AUC of 0.700 at iteration 11. A similar pattern appears for Qwen2, where the RF reaches its best ROC-AUC of 0.720 at iteration 8 and the language model peaks at 0.705 at iteration 12. The RF reaches its peak before the language models in both cases, which implies that RF performance responds more quickly to representation updates than the language model itself. As shown in Figure~\ref{fig:dynamics}(a), the RF trajectory exhibits moderate fluctuations during training, while the language model improves more gradually and remains stable. Both model components continue to benefit from reciprocal feedback beyond the initial iterations, although the magnitude of improvement decreases as convergence is approached.

Convergence behavior differs more noticeably on the Breast Cancer dataset. Qwen2 reaches near-peak performance after the first reciprocal update and satisfies the early-stopping criterion after only seven iterations. In contrast, CBERT continues improving until iteration 18 before converging. Despite the difference in convergence speed, both models ultimately achieve near-ceiling performance, with final ROC-AUC values exceeding 0.998.

The Diabetes dataset exhibits the fastest stabilization after the initial performance gains. Both CBERT and Qwen2 reach their strongest RF performance within the first few iterations and satisfy the convergence criterion shortly thereafter. The smaller ROC-AUC improvements observed on this dataset are consistent with the strong standalone baselines and large sample size, which reduce the opportunity for substantial gains. The rapid stabilization suggests that most useful information exchange occurs during the early stages of co-training, after which the early-stopping mechanism identifies the point of diminishing returns.

Figure \ref{fig:dynamics}(b) presents the evolution of the PPO reward during training. Average rewards increase rapidly during the initial stages and then plateau as training progresses. This stabilization reveals that the language model progressively aligns with both the supervised objective and RF-derived feedback. The absence of large fluctuations or reward collapse means that the hybrid reward formulation provides a stable learning signal throughout optimization.

Additionally, RCT converges within a modest number of iterations across all datasets and encoder architectures. Most performance gains are achieved during the first 5--10 iterations, after which improvements diminish and the early-stopping criterion terminates training automatically. These results suggest that the proposed framework remains computationally practical despite its iterative nature.

\begin{table}[!t]
\centering
\small
 \setlength{\tabcolsep}{5pt}
\begin{threeparttable}
\caption{
Ablation study on the MS-Relapse dataset
}
\label{tab:ablation}
\begin{tabular}{lll}
\toprule
\textbf{Component} & \textbf{Variant} & \textbf{ROC-AUC ($\Delta$)} \\
\midrule

\multicolumn{3}{c}{\textit{LM-focused ablations}} \\
\midrule

Training Scheme
& Iterative (Full RCT)
& \textbf{0.700} \\
& One reciprocal iteration
& 0.694 (-0.006) \\

Optimization
& PPO
& 0.694 \\
& REINFORCE
& 0.692 (-0.002) \\

Reward
& Hybrid ($R_\text{RF}+R_\text{acc}$)
& 0.606 \\
& RF reward only
& 0.593 (-0.013) \\
& Accuracy reward only
& 0.595 (-0.011) \\

Entropy $\beta$
& 0.05
& 0.677 \\
& 0.10
& 0.676 (-0.001) \\

\midrule

\multicolumn{3}{c}{\textit{RF-focused ablations}} \\
\midrule

Training Scheme
& Iterative (Full RCT)
& \textbf{0.721} \\
& Frozen LM 
& 0.708 (-0.013) \\
& One reciprocal iteration
& 0.705 (-0.016) \\

PCA Dimension
& 10
& 0.704 (-0.017) \\
& 20
& 0.705 (-0.016) \\

\bottomrule
\end{tabular}

\begin{tablenotes}[flushleft]
\small
\item Bold values denote the full RCT model. $\Delta$ denotes the absolute ROC-AUC difference from the full RCT configuration.
\end{tablenotes}
\end{threeparttable}
\end{table}

\begin{figure*}[!t]
    \centering
\includegraphics[ trim = 0 0 0 0, clip, scale=0.23]
{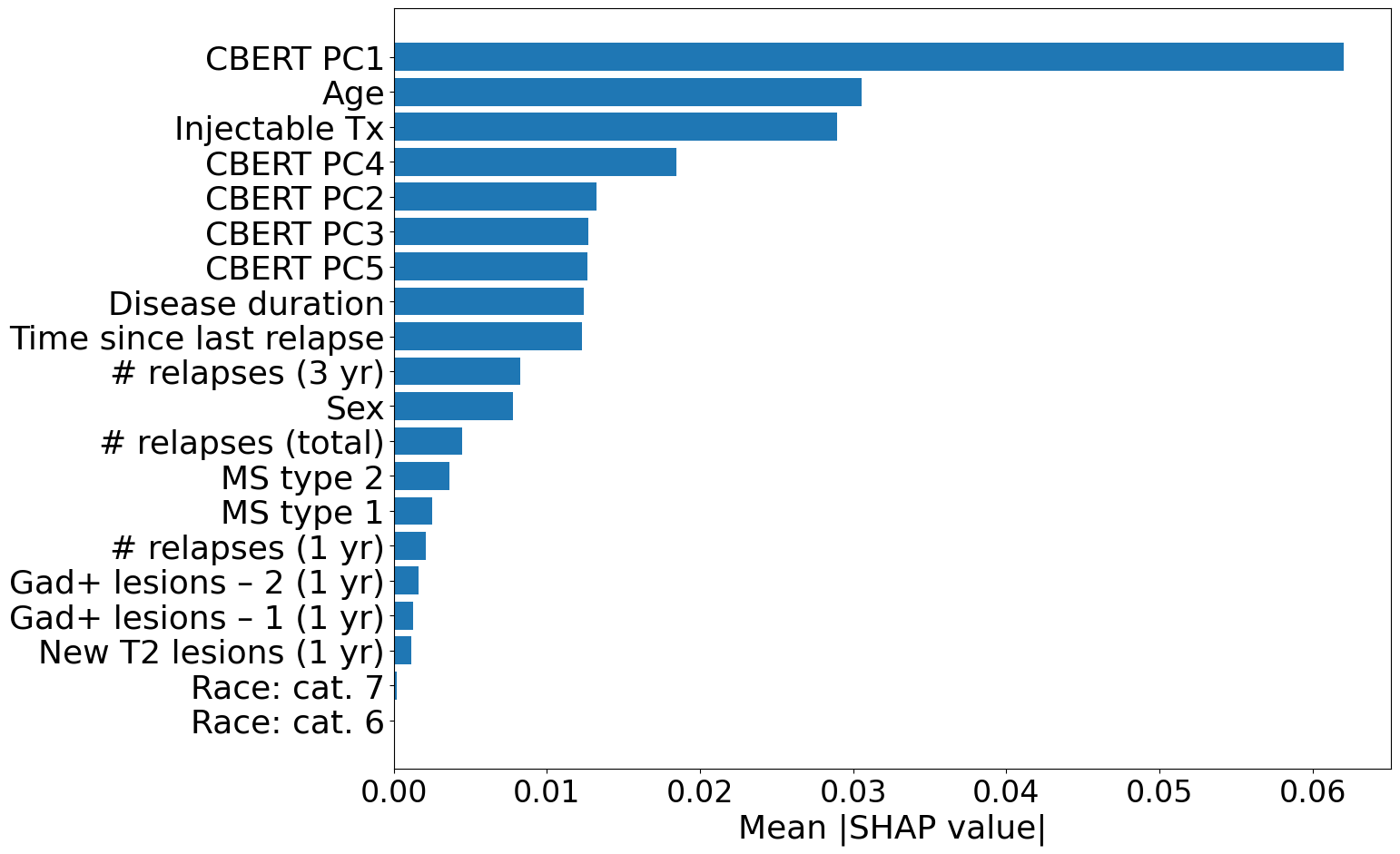}
\includegraphics[ trim = 0 0 0 0, clip, scale=0.23]
{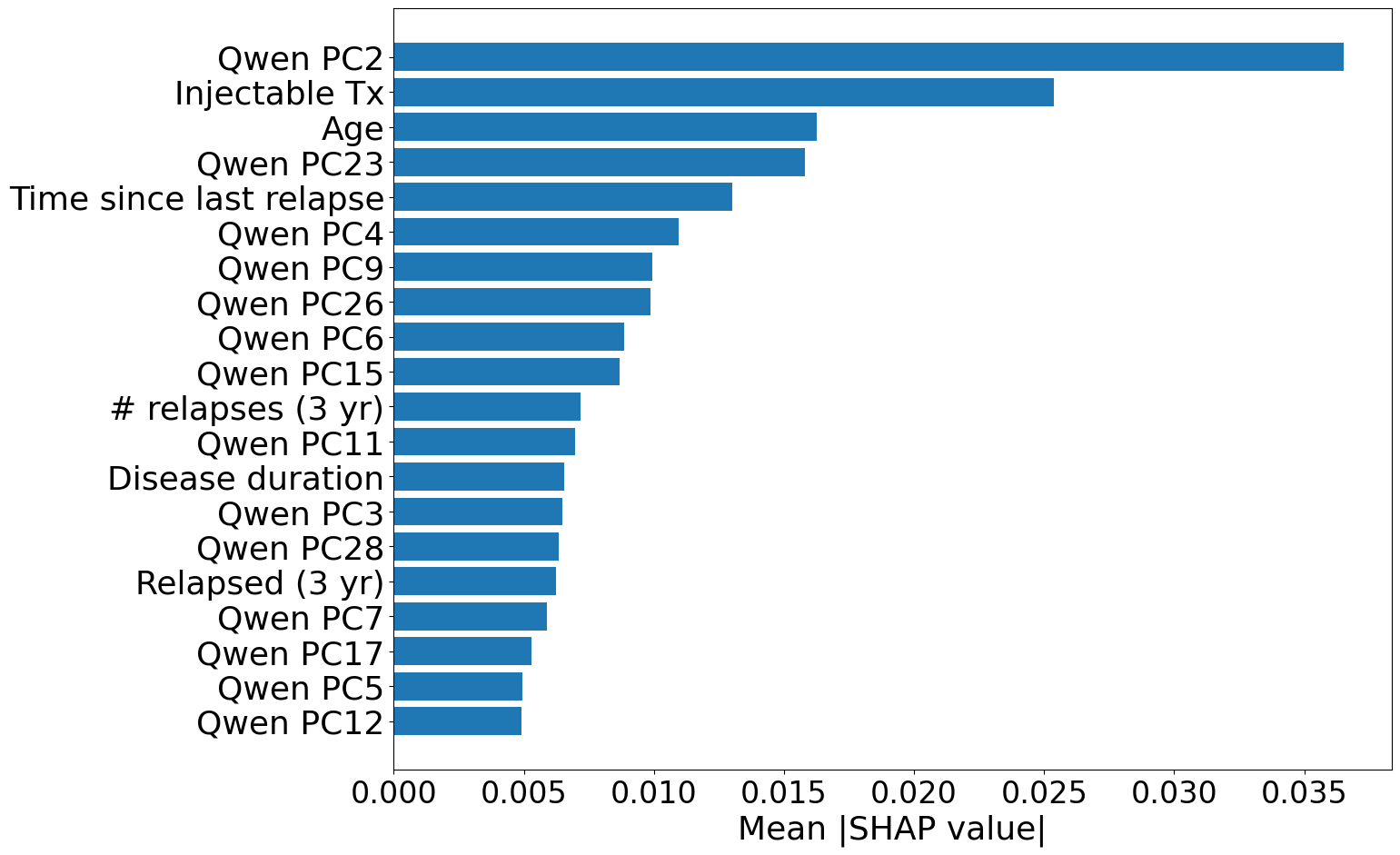}

    \caption{Top-20 RF features on the MS-Relapse dataset ranked by mean absolute SHAP value. LM-derived PCA components and original clinical variables are reported. Left: CBERT. Right: Qwen2.}
    \label{fig:shap}
\end{figure*}
\subsection{Ablation Analysis}

Due to computational cost, ablation studies were conducted with CBERT on the MS-Relapse dataset only. The experiments evaluate the contribution of key components of the RCT framework (Table~\ref{tab:ablation}), including iterative refinement, reward formulation, optimization algorithm, and embedding dimensionality. All experiments employ the same training protocol and hyperparameters unless otherwise noted. 

\textbf{Iterative refinement.}
Iterative updates are compared with a one-reciprocal-iteration variant in which the RCT loop executes exactly one exchange: the LM undergoes one PPO stage using the initial RF as a fixed teacher, and the RF is subsequently retrained once using the resulting embeddings. A second baseline, denoted Frozen LM, trains the RF on tabular features augmented with PCA-reduced embeddings extracted from the post-LoRA LM checkpoint, without PPO updates or reciprocal feedback.

For the LM component, one reciprocal iteration reaches a ROC-AUC of 0.694, below the full RCT performance of 0.700. The convergence results in Table~\ref{tab:convergence} show that the benefit of continued iteration varies by dataset and model. 
For the RF component, Frozen LM achieves a ROC-AUC of 0.708, compared with 0.705 for one reciprocal iteration and 0.721 for the full RCT framework. Notably, one reciprocal iteration performs slightly below the Frozen LM baseline. This result implies that a single reciprocal update is not sufficient to realize the full benefit of reciprocal co-training. The Frozen LM baseline demonstrates that static LM representations alone improve RF performance relative to the standalone RF baseline. However, the additional gain achieved by the full RCT framework confirms that reciprocal adaptation provides gains beyond simple feature augmentation. The improvement from 0.708 to 0.721 suggests that iterative refinement allows the RF to benefit from progressively improved representations.

\textbf{Reward formulation.}
The hybrid reward $R = R_{\text{RF}} + \lambda R_{\text{acc}}$ is evaluated by ablating each component. Using only the RF-derived reward reduces ROC-AUC from 0.606 to 0.593, while using only the classification-accuracy reward reduces ROC-AUC to 0.595. These results provide evidence that both reward components contribute to performance, with the classification-accuracy reward providing a slightly larger contribution.

\textbf{Optimization algorithm and entropy regularization.}
PPO achieves a slightly higher ROC-AUC than REINFORCE (0.694 vs. 0.692). The entropy regularization coefficient has little effect, with $\beta = 0.05$ and $\beta = 0.10$ producing nearly identical ROC-AUC values.

\textbf{PCA dimensionality.}
For CBERT, the full RCT configuration uses PCA dimension 5. Increasing the dimensionality to 10 or 20 reduces RF performance, yielding ROC-AUC values of 0.704 and 0.705, respectively.

\subsection{Feature Importance Analysis}

SHAP analysis was performed on the final RF models trained under the full RCT framework to examine the contribution of LM-derived representations. Figure~\ref{fig:shap} presents the 20 most influential features ranked by mean absolute SHAP value for both model architectures.

For the CBERT-based model, all five retained PCA components appear among the top 20 features, with CBERT PC1 ranked first overall. For the Qwen2-based model, 14 of the top 20 features correspond to PCA components derived from LM embeddings, with Qwen PC2 ranked as the most influential feature overall. Age, injectable treatment status, disease duration, and relapse history also rank highly for both models, suggesting that the RF retains clinically relevant information alongside the LM-derived representations.

Across both architectures, the embeddings derived from the language models account for many of the most influential RF features, which demonstrates that the learned representations contribute to the prediction along with the original clinical variables. This finding is consistent with the Frozen LM ablation, which demonstrates that embedding-based augmentation improves RF performance. These results support that the RF benefits from the information captured by the learned representations in addition to the original tabular feature representation.

\section{Limitations and Ethical Considerations}

The current study has several limitations. First, the MS-Relapse dataset is derived from a single clinical center, which may limit generalizability to broader patient populations. Second, the experiments focus on binary prediction tasks, and evaluation on more complex clinical outcomes is left to future work. Third, reciprocal co-training introduces additional computational cost relative to conventional supervised learning pipelines because both the LM and RF components must be updated repeatedly throughout training. The current study evaluates reciprocal adaptation only with Random Forests. Although the framework is applicable to other non-differentiable learners, such as gradient-boosted trees and rule-based models, empirical validation of these settings remains future work.

The proposed framework is intended to support, rather than replace, 
clinical decision-making. Predictions generated by the model should be 
interpreted as risk estimates and should not serve as the sole basis 
for treatment decisions. Although the experiments use de-identified 
clinical and public health datasets, additional external validation is 
necessary before deployment in real-world healthcare settings. Model 
performance may vary across demographic subgroups, and systematic bias 
in the training data could propagate through both the language model 
and RF components. Additional evaluation across diverse patient 
populations is needed to ensure equitable and reliable clinical use.

\section{Conclusion}

This work introduces a reciprocal co-training framework that couples a language model encoder with a Random Forest classifier through alternating optimization, reinforcement learning, and a hybrid reward design. By reformulating tabular data into standardized textual representations, contextual language model embeddings augment the RF feature space, while calibrated RF probability estimates provide feedback signals that guide language model updates via PPO.

Experiments across one proprietary clinical dataset and two public benchmarks demonstrate that the RCT framework provides consistent benefits across datasets relative to standalone baselines. Evaluation with both a domain-adapted encoder (CBERT) and an instruction-tuned language model (Qwen2) shows that the framework generalizes across fundamentally different encoder types. Ablation analyses confirm the importance of iterative refinement, hybrid reward formulation, and dimensionality control. SHAP analysis and Frozen LM ablation reveal that LM-derived embeddings contribute to RF performance and account for many of the most influential RF features.

The proposed framework offers a general mechanism for coupling gradient-based language models with non-differentiable learners through reciprocal adaptation. Future work may explore longitudinal prediction, multi-class and survival analysis settings, alternative non-differentiable learners, and evaluation under cross-institutional distribution shift.

\bibliographystyle{IEEEtran}
\bibliography{references}

@inproceedings{hollmann2023tabpfn,
  title={TabPFN: A Transformer That Solves Small Tabular Classification Problems in a Second},
  author={Hollmann, Noah and M{\"u}ller, Samuel and Eggensperger, Katharina and Hutter, Frank},
  booktitle={International Conference on Learning Representations (ICLR)},
  year={2023}
}

@article{dinh2022tablm,
  title={TabLM: Few-shot Classification of Tabular Data with Large Language Models},
  author={Dinh, Hieu and Nguyen, Hoang and Nguyen, Duy and others},
  journal={arXiv preprint arXiv:2210.10723},
  year={2022}
}

@article{yang2024qwen2,
  title={Qwen2 Technical Report},
  author={Yang, An and Yang, Baosong and Hui, Binyuan and Zheng, Bo and Yu, Bowen and Zhou, Chang and Fan, Chenguang and Liu, Dayiheng and Huang, Fei and Wei, Haoran and others},
  journal={arXiv preprint arXiv:2407.10671},
  year={2024}
}

@inproceedings{alsentzer2019clinicalbert,
  title={Publicly Available Clinical BERT Embeddings},
  author={Alsentzer, Emily and Murphy, John and Boag, William and Weng, Wei-Hung and Jin, Di and Naumann, Tristan and McDermott, Matthew},
  booktitle={Proceedings of the 2nd Clinical Natural Language Processing Workshop},
  pages={72--78},
  year={2019},
  publisher={Association for Computational Linguistics},
  url={https://aclanthology.org/W19-1909/}
}

@article{github,
  title        = {{Reciprocal Co-Training Framework Implementation}},
  author={{Anonymous GitHub Repository}},
  year = {2026},
  url = {https://anonymous.4open.science/r/Reciprocal-Co-Training-RCT-Coupling-Gradient-Based-and-Non-Differentiable-Models-via-Reinforcemen-45DC/}
}

@inproceedings{madill2024prediction,
  title={Prediction of Annualized Relapse Rate at First Clinic Visit Among Patients with Multiple Sclerosis (P5-6.015)},
  author={Madill, Evan and Healy, Brian and Polgar-Turcsanyi, Mariann and Chitnis, Tanuja},
  booktitle={Neurology},
  volume={102},
  number={7\_supplement\_1},
  pages={6504},
  year={2024},
  organization={Lippincott Williams \& Wilkins Hagerstown, MD},
  url={https://www.neurology.org/doi/abs/10.1212/WNL.0000000000206512}
}

@article{huang2019clinicalbert,
  title={ClinicalBERT: Modeling clinical notes and predicting hospital readmission},
  author={Huang, Kexin and Altosaar, Jaan and Ranganath, Rajesh},
  journal={arXiv preprint arXiv:1904.05342},
  year={2019},
  url={https://arxiv.org/abs/1904.05342}
}

@inproceedings{gururangan2020don,
  title={Don’t Stop Pretraining: Adapt Language Models to Domains and Tasks},
  author={Gururangan, Suchin and Marasovi{\'c}, Ana and Swayamdipta, Swabha and
          Lo, Kyle and Beltagy, Iz and Downey, Doug and Smith, Noah A.},
  booktitle={Proceedings of ACL},
  year={2020}
}

@inproceedings{vaswani2017attention,
  title={Attention Is All You Need},
  author={Vaswani, Ashish and Shazeer, Noam and Parmar, Niki and 
          Uszkoreit, Jakob and Jones, Llion and Gomez, Aidan N. and 
          Kaiser, Lukasz and Polosukhin, Illia},
  booktitle={Advances in Neural Information Processing Systems},
  year={2017}
}

@inproceedings{brown2020language,
  title={Language Models are Few-Shot Learners},
  author={Brown, Tom B. and Mann, Benjamin and Ryder, Nick and 
          Subbiah, Melanie and Kaplan, Jared and Dhariwal, Prafulla and 
          Neelakantan, Arvind and Shyam, Pranav and Sastry, Girish and 
          Askell, Amanda and Agarwal, Sandhini and Herbert-Voss, Ariel and 
          Krueger, Gretchen and Henighan, Tom and Child, Rewon and 
          Ramesh, Aditya and Ziegler, Daniel and Wu, Jeffrey and 
          Winter, Clemens and Hesse, Christopher and Chen, Mark and 
          Sigler, Eric and Litwin, Mateusz and Gray, Scott and 
          Chess, Benjamin and Clark, Jack and Berner, Christopher and 
          McCandlish, Sam and Radford, Alec and Sutskever, Ilya and 
          Amodei, Dario},
  booktitle={Advances in Neural Information Processing Systems},
  year={2020}
}

@article{touvron2023llama,
  title={LLaMA: Open and Efficient Foundation Language Models},
  author={Touvron, Hugo and Lavril, Thibaut and Izacard, Gautier and 
          Martinet, Xavier and Lachaux, Marie-Anne and Lacroix, Timoth{\'e}e and 
          Rozi{\`e}re, Baptiste and Goyal, Naman and Hambro, Eric and 
          Azhar, Faisal and Rodriguez, Aurelien and Joulin, Armand and 
          Grave, Edouard and Lample, Guillaume},
  journal={arXiv preprint arXiv:2302.13971},
  year={2023}
}

@article{cdc2015brfss,
  title={Behavioral Risk Factor Surveillance System (BRFSS) Survey Data},
  author={{Centers for Disease Control and Prevention}},
  journal={U.S. Department of Health and Human Services, Centers for Disease Control and Prevention},
  year={2015},
  url={https://www.cdc.gov/brfss/annual_data/annual_2015.html}
}

@article{johnson2016mimiciii,
  title={MIMIC-III, a freely accessible critical care database},
  author={Johnson, Alistair E. W. and Pollard, Tom J. and Shen, Lu and Lehman, Li-wei H. and Feng, Mengling and Ghassemi, Mohammad and Moody, Benjamin and Szolovits, Peter and Celi, Leo Anthony and Mark, Roger G.},
  journal={Scientific Data},
  volume={3},
  pages={160035},
  year={2016},
  publisher={Nature Publishing Group},
  url={https://www.nature.com/articles/sdata201635}
}

@article{hu2021lora,
  title={LoRA: Low-Rank Adaptation of Large Language Models},
  author={Hu, Edward J. and Shen, Yelong and Wallis, Phillip and Allen-Zhu, Zeyuan and Li, Yuanzhi and Wang, Shean and Wang, Lu and Chen, Weizhu},
  journal={arXiv preprint arXiv:2106.09685},
  year={2021},
  url={https://arxiv.org/abs/2106.09685}
}

@article{schulman2017ppo,
  title={Proximal Policy Optimization Algorithms},
  author={Schulman, John and Wolski, Filip and Dhariwal, Prafulla and Radford, Alec and Klimov, Oleg},
  journal={arXiv preprint arXiv:1707.06347},
  year={2017},
  url={https://arxiv.org/abs/1707.06347}
}

@article{gauthier2006model,
  title={A model for the comprehensive investigation of a chronic autoimmune disease: the multiple sclerosis CLIMB study},
  author={Gauthier, Susan A and Glanz, Bonnie I and Mandel, Micha and Weiner, Howard L},
  journal={Autoimmunity reviews},
  volume={5},
  number={8},
  pages={532--536},
  year={2006},
  publisher={Elsevier}
}

@misc{asuncion2007uci,
  title={UCI Machine Learning Repository},
  author={Asuncion, Arthur and Newman, David},
  year={2007},
  institution={University of California, Irvine, School of Information and Computer Sciences},
  url={https://archive.ics.uci.edu}
}

@article{street1993nuclear,
  title={Nuclear feature extraction for breast tumor diagnosis},
  author={Street, W. Nick and Wolberg, William H. and Mangasarian, Olvi L.},
  journal={Biomedical Image Processing and Biomedical Visualization},
  volume={1905},
  pages={861--870},
  year={1993},
  publisher={SPIE},
  url={https://doi.org/10.1117/12.148698}
}

@inproceedings{gorishniy2021revisiting,
  title={Revisiting Deep Learning Models for Tabular Data},
  author={Gorishniy, Yury and Rubachev, Ivan and Khrulkov, Valentin and Babenko, Artem},
  booktitle={Advances in Neural Information Processing Systems (NeurIPS)},
  year={2021},
  url={https://arxiv.org/abs/2106.11959}
}

@article{breiman2001random,
  title={Random Forests},
  author={Breiman, Leo},
  journal={Machine Learning},
  volume={45},
  number={1},
  pages={5--32},
  year={2001},
  publisher={Springer},
  url={https://link.springer.com/article/10.1023/A:1010933404324}
}

@article{grinsztajn2022why,
  title={Why do tree-based models still outperform deep learning on tabular data?},
  author={Grinsztajn, L{\'e}o and Oyallon, Edouard and Varoquaux, Ga{\"e}l},
  journal={Advances in Neural Information Processing Systems},
  volume={35},
  pages={507--520},
  year={2022},
  url={https://arxiv.org/abs/2207.08815}
}

@article{wolpert1992stacked,
  title={Stacked Generalization},
  author={Wolpert, David H.},
  journal={Neural Networks},
  volume={5},
  number={2},
  pages={241--259},
  year={1992},
  publisher={Elsevier},
  url={https://www.sciencedirect.com/science/article/pii/S0893608005800231}
}

@article{kablan2023evaluation,
  title={Evaluation of stacked ensemble model performance to predict clinical outcomes: A COVID-19 study},
  author={Kablan, Rianne and Miller, Hunter A and Suliman, Sally and Frieboes, Hermann B},
  journal={International Journal of Medical Informatics},
  volume={175},
  pages={105090},
  year={2023},
  publisher={Elsevier}
}

@article{alzubaidi2023stacking,
  title={Towards a stacking ensemble model for predicting diabetes mellitus using combination of machine learning techniques},
  author={Alzubaidi, Abdulaziz A. and Halawani, Sami M. and Jarrah, Mutasem},
  journal={International Journal of Advanced Computer Science and Applications},
  year={2023}
}

@article{akbar2022covid,
  title={COVID-19 detection using optimized AlexNet convolutional neural network with random forest classifier},
  author={Akbar, S. B. and others},
  journal={Computational Intelligence and Neuroscience},
  year={2022}
}

@inproceedings{chen2016xgboost,
  title={XGBoost: A scalable tree boosting system},
  author={Chen, Tianqi and Guestrin, Carlos},
  booktitle={Proceedings of the 22nd ACM SIGKDD International Conference on Knowledge Discovery and Data Mining},
  pages={785--794},
  year={2016}
}

@inproceedings{ouyang2022training,
  title={Training language models to follow instructions with human feedback},
  author={Ouyang, Long and Wu, Jeffrey and Jiang, Xu and Almeida, Diogo and Wainwright, Carroll L. and Mishkin, Pamela and Zhang, Chong and Agarwal, Sandhini and Slama, Katarina and Ray, Alex and others},
  booktitle={Advances in Neural Information Processing Systems},
  volume={35},
  pages={27730--27744},
  year={2022}
}

@inproceedings{ijcai2025p687,
  title     = {Latte: Transfering LLMs' Latent-level Knowledge for Few-shot Tabular Learning},
  author    = {Shi, Ruxue and Gu, Hengrui and Ye, Hangting and Dai, Yiwei and Shen, Xu and Wang, Xin},
  booktitle = {Proceedings of the Thirty-Fourth International Joint Conference on
               Artificial Intelligence, {IJCAI-25}},
  publisher = {International Joint Conferences on Artificial Intelligence Organization},
  editor    = {James Kwok},
  pages     = {6173--6181},
  year      = {2025},
  month     = {8},
  note      = {Main Track},
  doi       = {10.24963/ijcai.2025/687},
  url       = {https://doi.org/10.24963/ijcai.2025/687},
}

@phdthesis{zhan2023precision,
  author = {Zhan, Geng},
  title  = {Precision Monitoring for Disease Progression in Patients with Multiple Sclerosis: A Deep Learning Approach},
  school = {Dissertation},
  year   = {2023}
}

@article{chang2022detecting,
  author  = {Chang, J. Z. and others},
  title   = {Detecting multiple sclerosis disease activity and progression in progress notes from electronic medical records using natural language processing and machine learning},
  journal = {medRxiv},
  year    = {2022},
  doi     = {10.1101/2022.10.11.22280951}
}

\end{document}